\documentclass{article} 
\usepackage{iclr2027_conference,times}
\iclrfinalcopy 

\usepackage{amsmath,amsfonts,bm}

\def\eqref#1{equation~\ref{#1}}

\def\1{\bm{1}}

\DeclareMathAlphabet{\mathsfit}{\encodingdefault}{\sfdefault}{m}{sl}
\SetMathAlphabet{\mathsfit}{bold}{\encodingdefault}{\sfdefault}{bx}{n}

\usepackage{hyperref}
\usepackage{url}
\usepackage[utf8]{inputenc} 
\usepackage[T1]{fontenc}    
\usepackage{tabularx}
\usepackage{booktabs}       
\usepackage{amsfonts}       
\usepackage{nicefrac}       
\usepackage{microtype}      
\usepackage{xcolor}         
\usepackage{amsmath}
\usepackage{pifont} 
\usepackage{placeins}
\usepackage{float}
\usepackage{graphicx}
\usepackage[normalem]{ulem}
\usepackage{multirow}
\usepackage{wrapfig}
\usepackage[percent]{overpic}
\usepackage{needspace}
\newcommand{\xhdr}[1]{\vspace{2mm}\noindent{{\bf #1.}}}

\newcommand{\kenc}{knowledge encoding}
\newcommand{\Kenc}{Knowledge encoding}
\newcommand{\KEnc}{Knowledge Encoding}
\newcommand{\kext}{knowledge externalization}

\newcommand{\KExt}{Knowledge Externalization}

\title{
Playing Dumb: How a Chess Transformer \\Simulates Users Across Skill Levels 
}

\author{%
  Difan Jiao \quad
  George Eilender \quad
  Zhenwei Tang \quad
  Ashton Anderson \\[6pt]
  {\normalfont\small Department of Computer Science, University of Toronto} \\[2pt]
  {\normalfont\small $^{*}$\,Contact: \texttt{\{difanjiao, ashton\}@cs.toronto.edu}}
}

\begin{document}

\maketitle
\pagestyle{fancy}
\lhead{Preprint. Under review.}

\begin{abstract}

Generative models increasingly simulate users at different skill levels in education and AI evaluation. Yet it remains unclear whether lower-skill behavior reflects less internal knowledge or a different use of retained knowledge. This distinction is difficult to study in natural language, where proficiency and relevant knowledge are hard to measure. In this work, we study Maia-2, a chess transformer that predicts human moves conditioned on player skill, using quantitative skill ratings and deterministically labeled chess concepts. We hold board positions and model weights fixed and analyze positions where skill conditioning changes the prediction from a blunder to the best move. First, we measure encoded knowledge by training probes to predict the presence of 172 chess concepts from Maia-2's internal states. Probe accuracy remains nearly constant across skill conditions, and probes trained at one level transfer accurately to the others. To test whether stronger decisions can be recovered from this retained knowledge, we fine-tune only Maia-2's final output layer, raising best-move accuracy in the lowest skill band from 14.4\% to 82.8\%. Training this layer only on highest-skill representations retains most of the improvement at lower levels, showing that recovery does not require direct training on lower-skill states. Steering concept-relevant features learned by a sparse autoencoder also improves lower-skill decisions while leaving probe-measured concept encoding practically unchanged. These results support an account in which skill conditioning changes how retained concept knowledge influences move selection, showing that weaker behavior can understate the knowledge available within a model. Code is available at \url{https://github.com/CSSLab/maia2-skill-adaptation}.

\end{abstract}

\section{Introduction}

Modern generative models can adapt their outputs to different skill levels~\citep{brown2020language, wei2022emergent, li2024benchmarking}. Users can prompt large language models (LLMs) with instructions such as ``Act as if you were a world-class programmer'' or ``Pretend you don't know much about trigonometry''. Matching human behavior at different skill levels is also an explicit objective in specialized applications. Recent work has explored using LLMs to simulate learners at different proficiency levels in education~\citep{lu2024generative, jin2025teachtune, benedetto2024using}, while Maia-2 predicts human chess moves calibrated to player ratings~\citep{tang2024maia2}. Across these settings, skill conditioning changes a model's behavior while its weights remain fixed.

\begin{wrapfigure}{r}{0.45\linewidth}
    \centering
    \includegraphics[width=\linewidth,trim=0 5pt 0 0,clip]{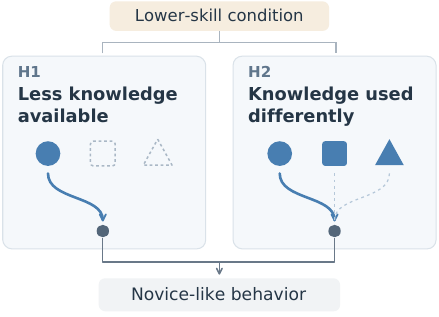}
    \vspace{-15pt}
    \caption{Two hypotheses for novice-like behavior: less available knowledge (H1) or a different use of retained knowledge (H2).}
    \label{fig:teaser_figure1}
    \vspace{-1pt}
\end{wrapfigure}

When a model is conditioned to behave like a novice, it may have less knowledge available or use what it knows differently (Figure~\ref{fig:teaser_figure1}). The distinction matters when behavior is used to infer what a model knows. In AI safety, oversight often relies on evaluations of what a model can do. Yet models can be prompted or fine-tuned to underperform on evaluations~\citep{van2024ai}, hide capabilities behind a password while imitating a weaker model~\citep{greenblatt2024stress}, or conceal their abilities while pursuing goals~\citep{meinke2024frontier}. An overseer relying on these outputs may therefore underestimate what the model could do under different conditions~\mbox{\citep{burns2023weak, anthropic2025recommendations}}. In education, the same issue arises when a simulated learner is used to model what a student knows~\citep{corbett1994knowledge}. Novice-like errors alone do not establish that the simulation reflects the student's knowledge~\citep{yuan2026towards}.

Previous work on skill conditioning in LLMs has primarily focused on controlling their behavior and assessing whether it matches a specified proficiency level~\citep{benedetto2024using, guollms}. However, how conditioning produces these behavioral differences remains unclear. Distinguishing changes in available knowledge from changes in its use is difficult in natural language, where both proficiency and relevant knowledge are hard to define and measure. Fluent role-play complicates this distinction: a novice-like response may draw on knowledge the simulated learner lacks~\citep{yuan2026towards}.

To resolve these challenges, we study skill-level adaptation in chess, where player ratings quantify skill, engines evaluate move quality, and concepts such as threats to individual pieces can be precisely defined from the board position~\citep{hxim2024stockfish,leelachesszero2024,pythonchess2024}. Moreover, models like Maia-2 explicitly condition their move predictions on player skill, allowing us to compare their decisions on the same board across ratings while keeping its weights fixed~\citep{tang2024maia2}. Maia-2 is thus an ideal simulated user: conditioned on a player's rating, it is trained to reproduce that player's moves, including their mistakes. Existing methods for interpreting chess models also provide tools for examining which concepts a model represents and testing how those concepts influence its moves~\citep{mcgrath2022acquisition,karvonen2024measuring,jenner2024evidence}.

\begin{figure}[t!]
    \centering
    \includegraphics[width=0.94\textwidth]{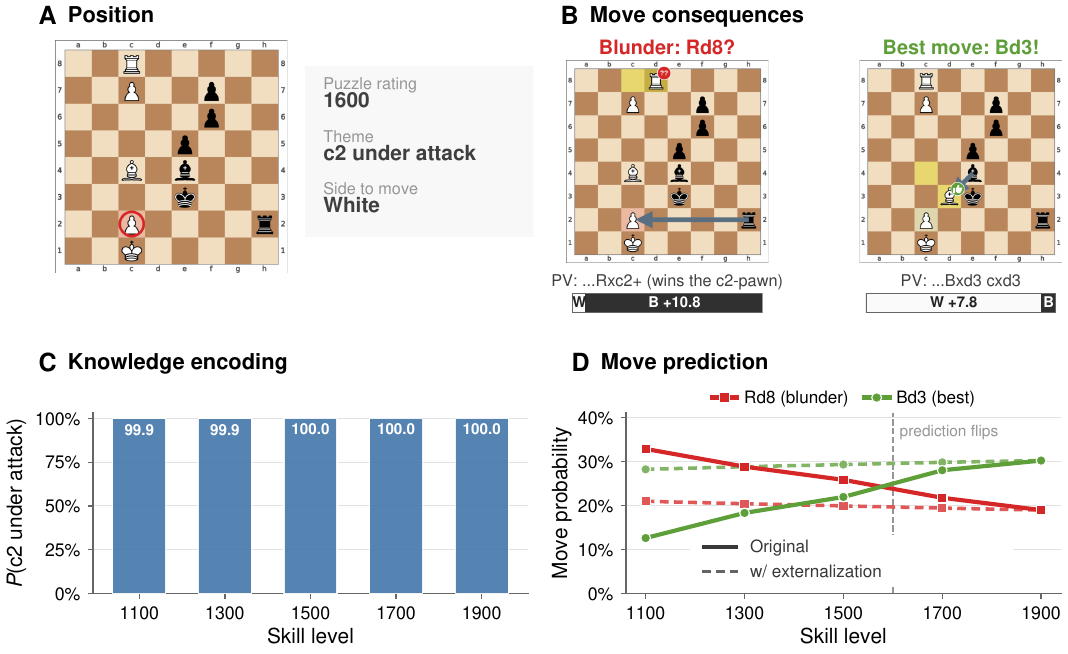}
    \vspace{-7pt}
    \caption{An illustrative example of Maia-2's skill-level adaptation. (A) The puzzle position and its relevant theme. (B) The consequences of the blunder and best move. (C) Probe-measured encoding of the relevant threat. (D) Maia-2's probabilities for the two moves across skill levels.}
    \label{fig:teaser}
\end{figure}

In this work, we study how Maia-2 adapts its play across skill levels. We focus on positions where the model makes a blunder\footnote{In chess, a blunder is a serious mistake that substantially reduces a player's chance of winning.} under lower-skill conditions but selects the best move under higher-skill conditions~\citep{tang2024maia2}, as illustrated in Figure~\ref{fig:teaser} (B). We first ask whether weaker play reflects less knowledge of relevant chess concepts. To make this question measurable, we examine 172 human-interpretable chess concepts using linear probes. These simple classifiers predict whether a concept is present on the board using the representations Maia-2 computes while processing that position~\citep{mcgrath2022acquisition}. We use probe accuracy to measure \kenc{}, or how consistently these concepts are represented under each skill condition. \textbf{Hypothesis 1 (H1): skill-dependent \kenc{}} predicts that these concepts should be less consistently represented at lower skill levels. However, probe accuracy remains nearly constant from novice to expert conditions, even as move quality changes sharply (Figure~\ref{fig:teaser} (C)). Moreover, probes trained at one skill level remain accurate at other levels without retraining, showing that this stability does not depend on fitting a separate classifier to each condition. Together, these results do not support a loss of the measured concept knowledge as the explanation for weaker play.

We therefore test \textbf{Hypothesis 2 (H2): skill-dependent \kext{}}, which proposes that skill conditioning changes how Maia-2 uses retained concept knowledge to select moves. We ask whether targeted interventions can recover high-skill decisions from lower-skill representations while preserving measured \kenc{}. We first fine-tune a separate policy head for each target concept, changing only the output layer that maps Maia-2's representations to move predictions while leaving the model's encoded concept knowledge unchanged. On held-out positions, this raises best-move accuracy in the lowest skill band from 14.4\% to 82.8\%, closing most of the behavioral gap. When applied to other concepts, these heads lose much of this improvement, indicating concept specialization. As a complementary test, we use sparse autoencoder (SAE) feature steering to amplify internal features associated with the target concepts~\citep{templeton2024scaling}. This partially closes the behavioral gap while leaving probe-measured \kenc{} practically unchanged. Together, these results support H2 by showing that stronger decisions can be recovered from lower-skill representations while measured \kenc{} remains stable.

To our knowledge, this is the first separation of knowledge encoding and knowledge use in a transformer-based generative model. By characterizing \kenc{} and \kext{}, we show that Maia-2's measured \kenc{} remains stable as its skill-conditioned move decisions change, while targeted interventions recover high-skill decisions from lower-skill states. Together, this framework provides a foundation for testing whether analogous capability-behavior gaps arise in other skill-conditioned transformers, including LLMs, when both skill levels and domain knowledge can be measured cleanly.


\FloatBarrier

\section{Related Work}


\xhdr{Skill conditioning and capability} LLMs can be conditioned to simulate learners at different proficiency levels~\citep{lu2024generative,jin2025teachtune,benedetto2024using}. Fidelity varies across models and proficiency scales, and simulated novices may retain expert reasoning~\citep{benedetto2024using,guollms}; valid student simulation therefore requires matching the learner's knowledge state~\citep{yuan2026towards}. In AI evaluation, models can also be prompted or fine-tuned to underperform selectively or hide capabilities while imitating weaker models~\citep{van2024ai,li2025covertly,griefalbert2026emulate,greenblatt2024stress,taylor2025auditing}. These studies show that behavior can misrepresent knowledge or capability, but leave unclear how a fixed model changes its internal computations across skill conditions.

\xhdr{Skill acquisition and localization} Other work studies how skills emerge during training and how training data shapes them~\mbox{\citep{arora2023theory,liu2025physics,chen2024skill}}. \mbox{Interpretability studies} also identify neurons and circuits associated with specific task capabilities~\mbox{\citep{wang2022finding,gurnee2023finding,dunefsky2024transcoders}}. These studies examine how capabilities are acquired or where they are represented, rather than how their use varies across skill conditions.

\xhdr{Human--AI alignment in chess} Because superhuman engines do not necessarily predict human decisions, the Maia models instead optimize for human move prediction~\citep{mcilroy2020aligning,tang2025imitate}. Maia-2 extends this approach with a single skill-conditioned transformer that models human moves across player ratings~\citep{tang2024maia2}. Related work studies strong human-like play and skill calibration through search, as well as unified chess modeling~\citep{jacob2022modeling,zhang2024human,monroe2026chessformer}.

\xhdr{Mechanistic understanding of chess models} Chess has become a model system for mechanistic interpretability. \citet{mcgrath2022acquisition} use linear probes to track human-interpretable concepts acquired by AlphaZero, while \citet{karvonen2024measuring} evaluate sparse autoencoders by recovering interpretable board-game features. In addition, \citet{jenner2024evidence} use activation patching to study learned look-ahead in Leela Chess Zero. We build on these tools but study a different question: how concept representations and their influence on decisions vary across skill conditions.

Our work examines how \kenc{} and \kext{} vary across skill conditions within the same pretrained model. We combine concept probes with targeted interventions to test whether lower-skill representations retain concept knowledge supporting stronger decisions. Appendix~\ref{app:additional_related} reviews further related work on chess engines and on methods for reading and intervening on internal representations.

\section{Preliminaries}

\subsection{Maia-2 as a Base Model}
\label{sec:maia2}

We use Maia-2~\citep{tang2024maia2}, a single transformer-based model trained to predict human chess moves across player skill levels. Its objective is to match human play at a given level, including mistakes. Chess skill is quantified by Elo ratings, with higher ratings indicating stronger play. Maia-2 groups ratings into 11 categories, from below 1100 to at least 2000, which we call \emph{skill conditions}.

The model takes a chess position $\mathcal{B}$, including the piece locations and whose turn it is, together with the rating categories of the player to move, $e_a$, and the opponent, $e_o$. We denote its top-ranked move by $\hat{a}=f_\theta(\mathcal{B},e_a,e_o)$, where $\theta$ denotes the model parameters. Throughout our experiments, we set $e_a=e_o=e$. We vary $e$ while holding the position and model weights fixed.


As shown in Figure~\ref{fig:maia2_skill} in Appendix~\ref{app:maia2}, a convolutional encoder extracts board features, which are converted into tokens for Transformer blocks. Each rating category is mapped to a learned vector, or skill embedding. These embeddings modify the attention queries in each block, allowing the same board features to be processed differently across skill conditions. Attention and feed-forward layers update the tokens' running representations, known as the residual stream. A linear policy head, shared across skill conditions, maps the final block's residual stream to the predicted chess move logits. We refer to the network producing the head's input as the backbone. This distinction lets us examine concept representations within the backbone and test how changing the policy head affects move predictions. Appendix~\ref{app:maia2} provides further architectural details.

\subsection{Blundered Transitional Dataset}
\label{sec:data}

Because Maia-2 is trained to model human play, it does not select the engine-optimal move in every position. Therefore, to study the mechanism behind a clear skill-dependent behavioral change, we restrict our analysis to positions where higher-skill conditioning reliably improves its decision.

Specifically, a \emph{transitional position} is one where Maia-2's prediction changes monotonically from a suboptimal move at lower skill conditions to the engine-selected best move at higher conditions, without reverting~\citep{tang2024maia2}. We further require the lowest-skill prediction to be a blunder, defined here as a move that reduces the estimated winning probability of the side to move by more than 10 percentage points relative to the best move. The resulting \textbf{Blundered Transitional Dataset} contains more than one million positions and isolates a clear behavioral gap induced by skill conditioning. See Appendix~\ref{app:btd} for more details.


\section{Does Skill Conditioning Change Knowledge Encoding?}
\label{sec:h1_knowledge_encoding}

Because Maia-2's move quality changes sharply with its conditioning skill level, one natural explanation is that skill conditioning changes what the model represents internally: concepts required for strong play may be encoded less consistently under lower-skill conditions. We refer to this account as \textbf{Hypothesis 1: skill-dependent knowledge encoding}. Under H1, relevant chess concepts should be less consistently represented under lower-skill conditions.

\subsection{Characterizing Knowledge Encoding in Chess}
\label{sec:definition}

Following \citet{mcgrath2022acquisition}, we characterize chess knowledge in terms of the human-interpretable chess concepts represented in a model's internal states. Whereas they study what concepts AlphaZero represents, when they emerge during training, and where they appear in the network, we hold Maia-2's weights fixed and ask how these internal concept representations change across conditioning skill levels. We refer to how consistently these concepts are represented under each skill condition as \kenc{}.

To make this question measurable, we construct an oracle of 172 binary chess concepts spanning square control, king and queen safety, tactical patterns, and positional structure. Because each concept assigns a deterministic binary label $c(\mathcal{B}) \in \{0,1\}$ to a board position, the oracle provides ground-truth labels for analyzing internal representations. Crucially, these concepts are relevant to Maia-2's skill-dependent behavior: 93.6\% of positions in the Blundered Transitional Dataset contain at least one oracle concept that lower-skill predictions fail to address correctly and higher-skill predictions resolve. Appendix~\ref{app:oracle} lists all concepts and their definitions.

\subsection{Measuring Knowledge Encoding across Skill Levels}
\label{sec:probing}

We measure \kenc{} by testing how accurately linear probes recover oracle concept labels from Maia-2's internal states~\citep{mcgrath2022acquisition,alain2016understanding}. Specifically, for a board position $\mathcal{B}$, let $\mathbf{z}(\mathcal{B}) \in \mathbb{R}^{d_{\mathrm{res}}}$ denote its average-pooled residual stream after a transformer block under a given skill condition. For every concept, transformer block, and skill level, we then train an $\ell_1$-regularized probe by minimizing

\vspace{-5pt}
\begin{equation}
\mathbf{w}^* = \arg\min_{\mathbf{w}} \sum_{i=1}^N \mathcal{L}(g(\mathbf{z}(\mathcal{B}_i)\mid\mathbf{w}), c(\mathcal{B}_i)) + \lambda\lVert\mathbf{w}\rVert_1.
\end{equation}
\vspace{-5pt}

Here, $g(\cdot\mid\mathbf{w})$ is the probe, $N$ is the number of training positions, $\mathcal{L}$ is binary cross-entropy, and $\lambda$ controls regularization. To isolate the effect of skill conditioning, we fix the board positions, vary only Maia-2's skill condition, and compare probe accuracy on held-out positions across all 11 skill levels. We probe the 1024-dimensional residual stream after each of the two transformer blocks. Appendix~\ref{app:ka_all_concepts} details sampling, regularization, and training.

\begin{figure*}[!t]
    \centering
    \includegraphics[width=0.94\textwidth,trim=0 8pt 0 0,clip]{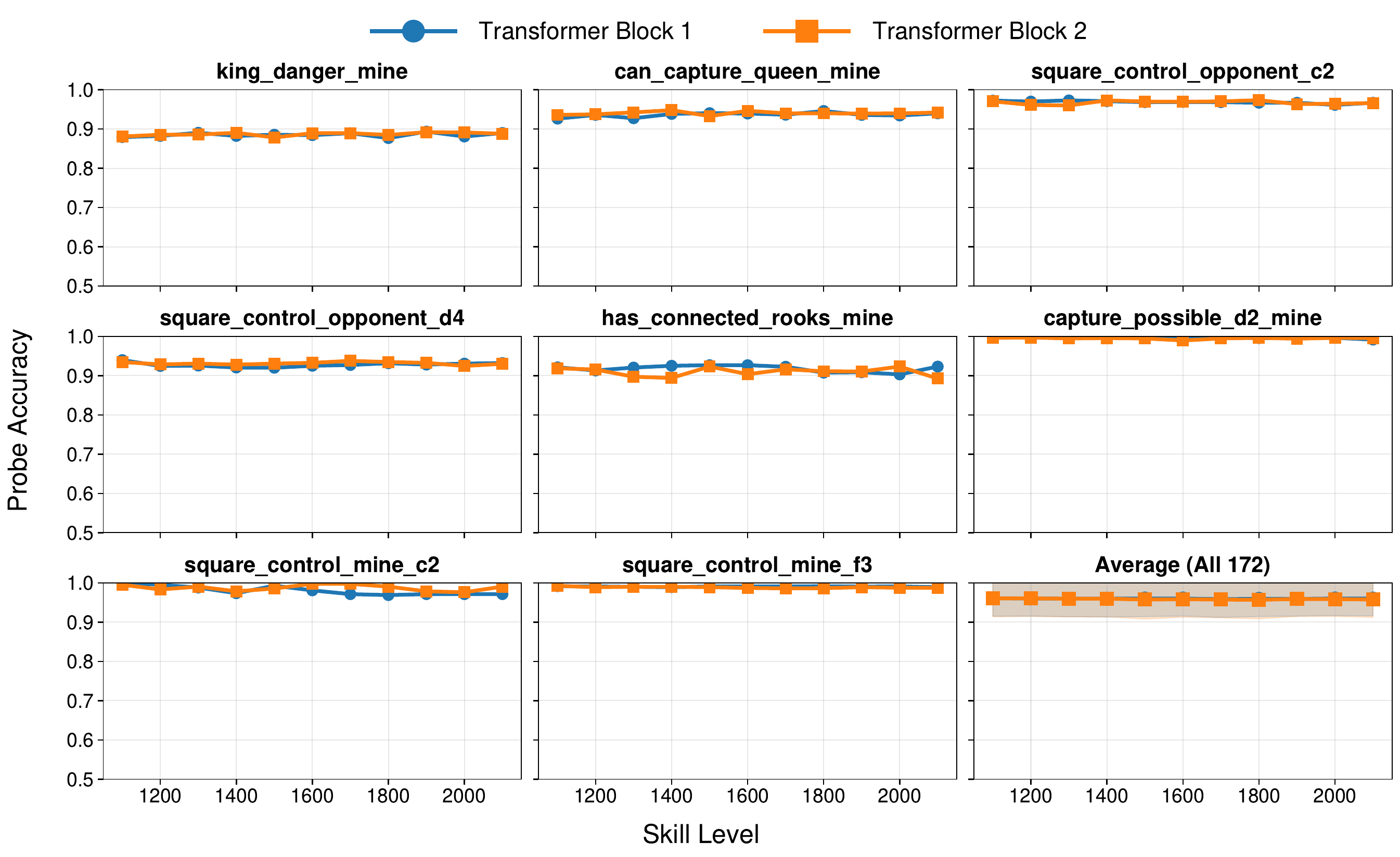}
    \vspace{-5pt}
    \caption{Linear probe accuracy across skill levels for eight representative concepts and the mean over all 172 oracle concepts.}
    \label{fig:probes}
    \vspace{4pt}

\end{figure*}

\subsection{Knowledge Encoding Remains Stable across Skill Levels}
\label{sec:probes}

Figure~\ref{fig:probes} shows results for eight representative concepts and the mean over all 172 concepts. Across the full oracle, mean test accuracy is 96.0\% after Transformer Block 1 and 95.9\% after Transformer Block 2. More importantly, the mean per-concept standard deviation across skill levels is only 0.57 percentage points for both blocks, and the same stability holds across all four concept categories. Thus, the near-constant mean does not conceal category-specific losses: lower-skill conditioning does not substantially reduce how consistently these chess concepts are represented. Appendix~\ref{app:ka_all_concepts} provides the complete per-concept results.

\xhdr{Cross-skill transfer} Within-skill accuracy alone leaves open whether each skill level requires a different linear readout. To test this alternative, we apply every source-skill probe, without refitting, to representations produced at every target skill level. Figure~\ref{fig:probe_transfer} in Appendix~\ref{app:probe_transfer} shows uniformly high accuracy across both matched and off-diagonal evaluations. Therefore, a largely shared linear readout remains usable across skill conditions, strengthening the evidence for stable \kenc{} despite modest decreases in transfer accuracy.

Together, the within-skill and cross-skill results do not support the loss of measured concept knowledge predicted by H1. Because Maia-2's move quality nevertheless varies sharply with skill, representational availability alone cannot explain the behavioral gap. We therefore next examine how the model translates its internal representations into move decisions.

\section{Can Stronger Decisions Be Recovered from Encoded Knowledge?}
\label{sec:h2_knowledge_externalization}
\label{sec:adjust_extern}

Section~\ref{sec:h1_knowledge_encoding} shows that Maia-2's measured \kenc{} remains stable across skill conditions, even though its move quality changes sharply. This mismatch creates a functional knowledge--action gap: concept information remains internally available, but lower-skill predictions often fail to act on it. A natural second explanation is therefore that skill conditioning changes how represented concepts influence move selection. We refer to this account as \textbf{Hypothesis 2: skill-dependent knowledge externalization}.


We characterize \kext{} as the extent to which concept-bearing internal representations are translated into decisions that act on the corresponding concepts. We test \kext{} through \emph{causal recoverability}: whether targeted interventions recover high-skill decisions from lower-skill representations while preserving measured \kenc{}. We first change only the policy head while keeping the backbone fixed. Under H2, a restricted readout intervention should substantially reduce the behavioral gap without changing the \kenc{} measured in Section~\ref{sec:h1_knowledge_encoding}.

This characterization concerns what the intervention can recover, not where Maia-2's original skill-adaptation mechanism is located. Specifically, Maia-2's Elo embeddings enter each transformer block, whereas its final policy head is shared across skill levels and receives no separate Elo input. Consequently, the policy head provides a controlled downstream intervention site, but success there does not imply that the pretrained model naturally implements skill adaptation only in that layer.


\subsection{Learning a High-Skill Readout with a Fixed Backbone}
\label{sec:pc_head_ft}

To test causal recoverability, we use \emph{Head SFT}: fine-tuning only Maia-2's policy head, the final linear layer mapping its 1024-dimensional backbone state to chess move logits and comprising only 8.2\% of the model's parameters. For each concept with sufficient qualifying positions, we initialize a separate head from the pretrained weights and train it on the corresponding concept-filtered subset of the Blundered Transitional Dataset. Specifically, let $\mathbf{z}_e(\mathcal{B}_i)$ denote the frozen backbone state for position $\mathcal{B}_i$ under skill condition $e$, and let $a_i^*$ denote the optimal move that Maia-2 already predicts correctly at its highest skill condition. We optimize

\vspace{-5pt}
\begin{equation}
    (\mathbf{W}_c^*, \mathbf{b}_c^*) =
    \arg\min_{\mathbf{W}_c,\mathbf{b}_c}
    \sum_{i=1}^{N_c}\sum_{e=e_{\min}}^{e_{\max}}
    \mathcal{L}_{\mathrm{CE}}\!\left(
      \mathbf{W}_c\mathbf{z}_e(\mathcal{B}_i)+\mathbf{b}_c,
      a_i^*\right).
\end{equation}
\vspace{-5pt}

Here, $\mathbf{W}_c$ and $\mathbf{b}_c$ are the parameters of the head for concept $c$, $N_c$ is the number of training positions for that concept, and $\mathcal{L}_{\mathrm{CE}}$ is cross-entropy loss. The bounds $e_{\min}$ and $e_{\max}$ denote the lowest and highest training skill conditions.

The target is distilled from Maia-2's own highest-skill prediction, but it is applied as supervision to the model's states at every skill level. The experiment therefore asks whether lower-skill states can be mapped to the same concept-appropriate decisions that Maia-2 produces at its highest skill level. Because the backbone remains frozen throughout training, every $\mathbf{z}_e(\mathcal{B})$ is identical before and after the intervention, and the measured \kenc{} is therefore preserved by construction. Appendix~\ref{app:ke_all} lists the filtering criteria and qualifying concepts.

\begin{table}[!t]
  \centering
  \renewcommand\arraystretch{1.0}
  \caption{Maia-2 performance on Blundered Transitional positions. STD is the sample standard deviation of five band means; other subscripts show SD across concepts.}
  \vspace{0.3cm}
  \setlength{\tabcolsep}{5pt}
    \begin{tabular}{p{3.3cm}|cccc}
    \toprule
    & Baseline & Steering (Random) & Steering (Feature) & Head SFT \\
    \midrule
    \multicolumn{5}{l}{\textbf{Best-Move Accuracy}} \\
    \quad $<1200$ & $0.144_{0.036}$ & $0.163_{0.042}$ & $0.427_{0.086}$ & $0.828_{0.090}$ \\
    \quad 1200--1399 & $0.443_{0.078}$ & $0.456_{0.059}$ & $0.573_{0.065}$ & $0.854_{0.078}$ \\
    \quad 1400--1599 & $0.621_{0.072}$ & $0.629_{0.051}$ & $0.698_{0.052}$ & $0.870_{0.070}$ \\
    \quad 1600--1799 & $0.779_{0.053}$ & $0.781_{0.036}$ & $0.819_{0.035}$ & $0.883_{0.068}$ \\
    \quad 1800--1999 & $0.894_{0.030}$ & $0.894_{0.020}$ & $0.907_{0.019}$ & $0.912_{0.059}$ \\
    \midrule
    \multicolumn{5}{l}{\textbf{Skill Gap}} \\
    \quad STD ($\downarrow$) & $0.295$ & $0.287$ & $0.191$ & $\boldsymbol{0.031}$ \\
    \quad TP ($\downarrow$) & $4.238_{0.518}$ & $4.024_{0.528}$ & $3.415_{0.780}$ & $\boldsymbol{1.318_{0.673}}$ \\
    \midrule
    \multicolumn{5}{l}{\textbf{Move Quality}} \\
    \quad CP Loss ($\downarrow$) & $225.3_{47.5}$ & $207.4_{53.2}$ & $161.1_{55.6}$ & $\boldsymbol{61.8_{44.3}}$ \\
    \quad WR Loss ($\downarrow$) & $0.150_{0.024}$ & $0.135_{0.027}$ & $0.094_{0.033}$ & $\boldsymbol{0.026_{0.014}}$ \\
    \quad Blunder Rate ($\downarrow$) & $52.5\%_{5.5\%}$ & $47.3\%_{7.4\%}$ & $32.1\%_{10.5\%}$ & $\boldsymbol{8.0\%_{4.2\%}}$ \\
    \quad Ignore Rate ($\downarrow$) & $93.7\%_{4.4\%}$ & $83.8\%_{9.8\%}$ & $48.6\%_{23.3\%}$ & $\boldsymbol{4.1\%_{4.6\%}}$ \\
    \bottomrule
    \end{tabular}%
  \label{tab:prediction_accuracy}
  \vspace{0.3cm}
\end{table}

\subsection{A Linear Readout Recovers Most of the Skill Gap}
\label{sec:validate_2}

To measure how well each head externalizes its target concept, we evaluate it on held-out positions from the corresponding concept-filtered Blundered Transitional Dataset. Our primary metric is \emph{Best-Move Accuracy}, the fraction of positions where the head predicts the optimal move used to construct the dataset. We report this accuracy in five 200-Elo bands to show how the intervention changes decisions across the skill spectrum. Because H2 concerns the gap across skill conditions, we summarize cross-skill consistency with two additional metrics. \emph{STD} is the sample standard deviation of the five reported band means. \emph{TP} is the average transition point at which a position is first predicted correctly and remains correct at all higher levels; TP $=0$ means every level is correct, TP $=10$ means every level is wrong, and lower values indicate a smaller skill gap. Finally, we evaluate whether the predicted moves improve in chess terms at the skill levels below each position's baseline transition point. \emph{CP Loss} measures the centipawn degradation caused by a move, while \emph{WR Loss} converts this degradation into a change in win probability.\footnote{We use Lichess server-side evaluations when available (typically above depth 30); otherwise, we run Stockfish at depth 20. We convert centipawns with $\mathrm{WR}(cp)=(1+e^{-0.00368\,cp})^{-1}$ as in~\citet{lichess2024accuracy}.} \emph{Blunder Rate} is the fraction of moves with WR Loss $\geq 0.10$. \emph{Ignore Rate} is the fraction of moves that fail the oracle's check for addressing the target concept. For a threat to a piece, the check asks whether the move relocates or defends it.

Table~\ref{tab:prediction_accuracy} (Head SFT column) reports the results across 71 qualifying concepts. Head SFT raises best-move accuracy in the lowest Elo band from 14.4\% to 82.8\%, while accuracy in the highest reported band increases from 89.4\% to 91.2\%. Consequently, STD falls from 0.295 to 0.031, and mean TP falls from 4.24 to 1.32. The intervention therefore closes the gap primarily by improving lower-skill decisions rather than trading away high-skill performance. The chess-specific metrics show the same pattern. Below the baseline transition point, Head SFT reduces Ignore Rate from 93.7\% to 4.1\% and Blunder Rate from 52.5\% to 8.0\%. It also reduces average WR Loss from 0.150 to 0.026 and CP Loss from 225 to 62. Thus, the recovered predictions both address the active concept and substantially improve move quality. Because Head SFT changes only the final mapping from fixed representations to moves, these results show that lower-skill states support substantially stronger decisions that address the target concepts. A linear head can only reweight what the frozen representation already contains, so the recovered decisions were available in the lower-skill states. Two other explanations remain: the head may simply be a stronger general policy, or it may have fit the lower-skill states it was trained on. The next two analyses test each.


\Needspace{16\baselineskip}
\begin{wrapfigure}{r}{0.48\linewidth}
    \centering
    \includegraphics[width=\linewidth,trim=0 6pt 0 0,clip]{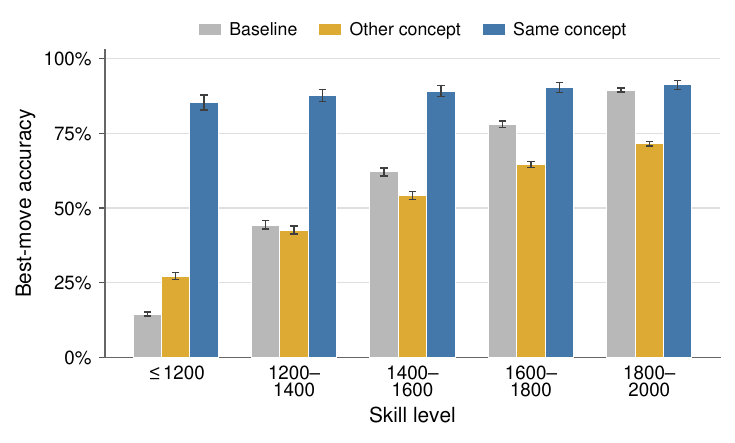}
    \vspace{-17pt}
    \caption{Best-move accuracy on target versus other concepts; error bars show 95\% bootstrap CIs across concepts.}
    \label{fig:transferability}
\end{wrapfigure}

\xhdr{Concept specialization} If policy head SFT simply learned a generic high-skill policy, a head trained on one concept should also improve positions governed by other concepts. To test this possibility, we independently rerun Head SFT for each concept, then apply every resulting head both to its own concept's test positions and to all other concepts' test positions. Figure~\ref{fig:transferability} shows that same-concept heads reproduce the large improvement above, whereas other-concept heads perform below the frozen baseline on average from the second skill band onward. These results indicate that the recovery is specific to the training concept and does not generalize as a uniformly stronger policy across concepts. This pattern is consistent with concept-specific \kext{}.

\Needspace{16\baselineskip}
\begin{wrapfigure}{r}{0.48\linewidth}
    \centering
    \includegraphics[width=\linewidth,trim=0 6pt 0 0,clip]{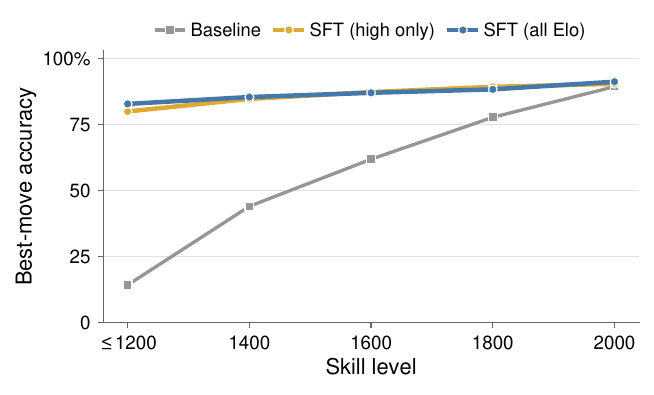}
    \vspace{-17pt}
    \caption{Best-move accuracy for policy heads trained on all skill conditions or only the highest-skill condition.}
    \label{fig:high_elo_only_transfer}
\end{wrapfigure}

\xhdr{Cross-skill readout transfer} Head SFT is trained on every position under all 11 skill conditions. Consequently, its success could still depend on direct supervised exposure to each lower-skill state. To isolate cross-skill transfer, we instead train and early-stop each concept-specific head using only the highest-skill representations, then evaluate the fixed head on held-out board positions under every skill condition. We repeat each high-skill training example 10 times. As Figure~\ref{fig:high_elo_only_transfer} shows, this high-skill-only readout raises best-move accuracy in the lowest skill band from 14.4\% to 80.0\% and remains within 2.9 percentage points of all-Elo Head SFT in every band. Thus, the mapping learned solely from high-skill states transfers directly to lower-skill states, and most of the recovery does not require supervised fitting on those states.

\subsection{Feature-level Intervention Partially Recovers the Skill Gap}
\label{sec:sae_steering}
\label{sec:sae_results}

Head SFT tests recovery by retraining the mapping from fixed representations to moves. Sparse autoencoders (SAEs) decompose internal activations into sparse features that can be manipulated to steer model outputs~\citep{templeton2024scaling,chalnev2024sae}. We use this approach to test whether amplifying concept-relevant features improves move selection while preserving measured \kenc{}.

Following work on JumpReLU SAEs and dictionary learning in chess~\citep{rajamanoharan2024jumping,karvonen2024measuring}, we train an SAE on Maia-2's internal activations and use oracle labels to select features most associated with each chess concept. At inference, we amplify these features and inject their decoded representation back into Maia-2. We select the number of features and intervention strength on training positions, allowing them to vary by concept and skill condition, and evaluate on held-out positions. \emph{Steering (Random)} uses the same procedure with randomly selected features. Appendix~\ref{app:feature_steering} details SAE training and steering.

Table~\ref{tab:prediction_accuracy} shows that concept-salient steering partially closes the behavioral gap. In the lowest skill band, best-move accuracy rises from 14.4\% to 42.7\%, compared with 16.3\% under random-feature steering. Correspondingly, mean TP falls from 4.24 to 3.42, whereas random-feature steering reaches 4.02. The chess-specific metrics show the same difference: feature steering reduces Ignore Rate from 93.7\% to 48.6\%, Blunder Rate from 52.5\% to 32.1\%, WR Loss from 0.150 to 0.094, and CP Loss from 225 to 161. Thus, Maia-2 produces stronger lower-skill decisions when we amplify concept-relevant internal features, even without retraining the model.

Finally, because feature steering directly intervenes on internal activations, its behavioral gains support H2 only if the measured \kenc{} remains stable. Figures~\ref{fig:sae_encoding_preservation} and~\ref{fig:post_steering_probes} in \mbox{Appendix~\ref{app:ke_all}} therefore evaluate probes trained \textit{before steering} on the intervened representations. After averaging over all skill conditions within each concept, mean probe accuracy changes by only 0.86 percentage points; the paired test supports practical equivalence within a margin of $\pm 1$ percentage point ($p=0.031$). Thus, SAE steering partially improves knowledge externalization while leaving probe-measured knowledge encoding practically unchanged on average.

\xhdr{Summary of evidence} Stable concept decoding and cross-skill probe transfer weigh against H1. Head SFT closes most of the behavioral gap with a fixed backbone, while SAE steering yields partial recovery with measured \kenc{} practically unchanged. These results primarily support H2, suggesting that Maia-2 retains the measured concept knowledge but uses it differently across skill conditions. The two intervention sites provide complementary evidence for this account. They do not establish H2 as the only mechanism of skill adaptation.


\FloatBarrier

\section{Discussion}

\subsection{Practical Implications}

\xhdr{Education} Maia-2's weaker decisions coexist with retained knowledge of the measured concepts. For simulations intended to model what a student knows, this illustrates why matching novice behavior alone cannot establish fidelity to the student's knowledge. Tutoring has a different objective: explanations and support can be adapted to a learner while the system retains expert knowledge. These roles require different validation criteria; neither knowledge retention nor knowledge loss is inherently preferable.

\xhdr{Capability evaluation} Recovering stronger decisions from lower-skill representations shows that performance under one condition can understate what a model can do with retained knowledge. This motivates capability assessments that vary conditioning contexts and test whether restricted interventions reveal stronger performance. For LLMs, such contexts include system instructions, role descriptions, and user prompts. Our experiments establish this recoverability in Maia-2; they do not establish that LLMs use the same mechanism or that weaker outputs reflect deliberate concealment of capabilities.

\subsection{Future Work}
\label{disc:future}

These results point to two directions for further study. First, extending the concept set to multi-move planning and long-term positional judgments would test whether the same pattern holds for knowledge that depends on possible continuations. Such concepts require validated measures beyond deterministic labels for the current board. Second, applying this analysis to LLMs would test its generality beyond chess. Unlike Maia-2's explicit rating conditions, prompting an LLM to imitate a proficiency level does not itself validate the resulting skill adaptation. Reliable proficiency conditions and independent knowledge measures would make that test informative.

\section{Conclusion}
\label{sec:conclusion}

We study skill-level adaptation in Maia-2 by distinguishing knowledge encoding from knowledge externalization. Comparing skill conditions on the same positions within a fixed pretrained model separates the effect of conditioning from differences in inputs or training. Across 172 static, deterministically labeled chess concepts, probe accuracy remains stable across skill conditions, and probes transfer accurately between them, even as move quality changes. Fine-tuning the policy head with a frozen backbone raises best-move accuracy in the lowest band from 14.4\% to 82.8\%, closing most of the behavioral gap. Most of this recovery remains when the head is trained only on highest-skill representations. SAE feature steering yields partial recovery while leaving probe-measured encoding practically unchanged on average. These results support an account in which skill conditioning changes how retained concept knowledge influences move selection. Combining probes with recovery tests thus provides a way to examine whether a model's weaker behavior reflects limited concept knowledge or a different use of the knowledge it retains.


\section*{Ethics Statement}
This work studies Maia-2, a model trained on human chess games, using publicly available Lichess data. Our experiments analyze model representations and move predictions and involve no new participant recruitment.

\section*{Reproducibility Statement}
Appendices~\ref{app:oracle}--\ref{app:ka_all_concepts} document the concepts, data construction, and probe training; Section~\ref{sec:pc_head_ft} and Appendix~\ref{app:feature_steering} describe the interventions. Code is available in the repository linked in the abstract.

\section*{AI Use Statement}
Generative AI assisted with writing, reference checks, figures, and formatting. The authors are responsible for the final content.

\clearpage
\bibliography{main}
\bibliographystyle{iclr2027_conference}

\clearpage
\appendix
\raggedbottom
\setcounter{topnumber}{4}
\setcounter{bottomnumber}{3}
\setcounter{totalnumber}{6}
\renewcommand{\topfraction}{0.95}
\renewcommand{\bottomfraction}{0.90}
\renewcommand{\textfraction}{0.05}
\renewcommand{\floatpagefraction}{0.90}
\setlength{\floatsep}{10pt plus 2pt minus 2pt}
\setlength{\textfloatsep}{5pt}
\setlength{\intextsep}{9pt plus 2pt minus 2pt}


\section{Extended Details for Maia-2}
\label{app:maia2}


\begin{wrapfigure}{r}{0.36\textwidth}
    \centering
    \includegraphics[width=\linewidth]{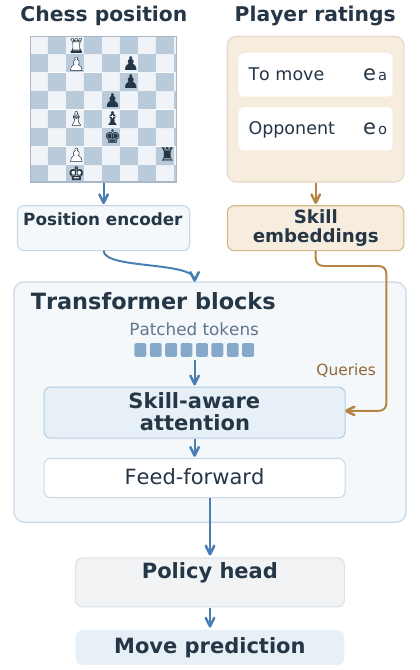}
    \caption{Maia-2 architecture. Skill-level embeddings condition Transformer attention before prediction.}
    \label{fig:maia2_skill}
    \label{fig:hermes}
\end{wrapfigure}

Maia-2 is a Transformer-based human move-prediction model conditioned on a chess position and the ratings of both players~\citep{tang2024maia2}. As Figure~\ref{fig:hermes} shows, it combines learned skill embeddings, a ResNet position encoder, and skill-aware attention before predicting the next move. We describe these components below and then explain why their combination provides a controlled setting for our study.

\subsection{Skill Embeddings}
Maia-2 discretizes player ratings into the 11 Elo categories defined in Section~\ref{sec:maia2} and maps each category to a learned embedding. Let $\mathbf{E} \in \mathbb{R}^{11 \times d_s}$ denote the embedding matrix. For the skill categories $a$ and $o$ of the active player and opponent, respectively, the model retrieves
$\mathbf{e}_a=\mathbf{E}[a]$ and $\mathbf{e}_o=\mathbf{E}[o]$.
Throughout our experiments, we set $a=o$ so that a single Elo category defines each skill condition.

\subsection{Position Encoding}
Each chess position is represented as an $18\times8\times8$ tensor: 12 channels encode piece locations, while the remaining channels encode the side to move, castling rights, and the en-passant square. A ResNet maps this tensor to eight spatial feature maps. Channel-wise patching then flattens each $8\times8$ feature map and projects it to a 1024-dimensional token, producing eight position tokens for the Transformer.

\WFclear
\subsection{Skill-Aware Attention}
Each Transformer block applies skill-aware multi-head self-attention followed by a feed-forward update, with both updates entering the residual stream. Skill conditioning acts on the attention queries. At block $\ell$, the concatenated player embeddings are linearly projected and added to the position-derived queries:
\begin{equation}
    \mathbf{Q}_{\ell}
    = \mathbf{Q}_{\ell}^{\mathrm{pos}}
    + \operatorname{reshape}\!\left((\mathbf{e}_a \oplus \mathbf{e}_o)
    \mathbf{W}_{\ell}^{\mathrm{skill}}\right),
\end{equation}
where $\oplus$ denotes concatenation and the skill-dependent term is broadcast across the position tokens. Consequently, changing the Elo condition can alter how the same position features interact while leaving the board and model weights fixed. After two Transformer blocks, the tokens are normalized and mean-pooled; a final layer normalization produces the 1024-dimensional state used by the policy, auxiliary, and value heads.

\subsection{Why Maia-2 Is Well Suited to This Study}
Maia-2 provides the three properties needed for our analysis. First, skill is an explicit, ordered conditioning variable within a fixed model. Therefore, we can hold the chess position and model weights constant while changing only the skill condition. Second, the conditions correspond to a quantitative human skill scale and produce meaningfully different move behavior: move quality generally improves with Elo, and many positions transition from lower-skill mistakes to higher-skill optimal moves. These positions provide the high-contrast testbed formalized in Section~\ref{sec:data}. Finally, access to the model's internal states and parameters lets us compare concept representations under the same conditions and apply restricted interventions without retraining the full model. Together, these properties make Maia-2 a well-instrumented setting for studying how skill conditioning changes behavior.

\section{Experimental Details and Extended Results}

\subsection{Oracle Chess Concept Definitions}
\label{app:oracle}

Table~\ref{tab:oracle_concepts} lists all 172 oracle chess concepts used in this work. As described in Section~\ref{sec:definition}, they span four categories. We add 128 square-control concepts and four king- and queen-safety concepts to capture fine-grained threats, while the 17 tactical-pattern and 23 positional-structure concepts follow the chess concepts studied by \citet{mcgrath2022acquisition}. All concepts are binary functions $c(\mathcal{B}) \in \{0,1\}$. We use \textbf{m} for the active player (``mine'') and \textbf{o} for the opponent; \texttt{[m\,|\,o]} denotes two separate concepts evaluated from the corresponding perspectives. Square-indexed concepts are instantiated independently for each listed square.

We restrict the oracle to atomic concepts with deterministic labels computed from the current board. Consequently, the labels require neither a learned annotator nor a search-based evaluation and remain fixed when we vary Maia-2's skill condition. This design supports controlled comparisons of the same concept across Elo conditions, but it does not measure dynamic abilities such as multi-move planning or search, as discussed in Section~\ref{disc:future}.

\begin{table}[H]
\centering
\caption{Glossary of the 172 binary oracle chess concepts.}
\label{tab:oracle_concepts}
\footnotesize
\renewcommand{\arraystretch}{0.88}
\begin{tabularx}{\linewidth}{@{}p{4.6cm} X@{}}
\toprule
\textbf{Concept names} & \textbf{Description} \\

\midrule
\multicolumn{2}{l}{\textit{Square Control --- Active Player \textnormal{(64 concepts)}}} \\
\midrule

\texttt{offensive\_threat\_\{sq\}} \newline sq $\in$ \{a1, \ldots, h8\} &
True if the opponent has a piece on square \texttt{sq} with strictly more active-player attackers than opponent defenders. Instantiated for each of the 64 squares. \\

\midrule
\multicolumn{2}{l}{\textit{Square Control --- Opponent \textnormal{(64 concepts)}}} \\
\midrule

\texttt{defensive\_threat\_\{sq\}} \newline sq $\in$ \{a1, \ldots, h8\} &
True if the active player has a piece on square \texttt{sq} with strictly more opponent attackers than own defenders. Instantiated for each of the 64 squares. \\

\midrule
\multicolumn{2}{l}{\textit{King and Queen Safety \textnormal{(4 concepts)}}} \\
\midrule

\texttt{king\_danger} [m\,|\,o] &
True if 5 or more squares adjacent to the side's king are attacked by the other side. \\

\texttt{can\_capture\_queen} [m\,|\,o] &
True if the side has a legal move that captures the other side's queen. \\

\midrule
\multicolumn{2}{l}{\textit{Tactical Patterns --- Forks \textnormal{(7 concepts)}}} \\
\midrule

\texttt{pawn\_fork} [m\,|\,o] &
True if an unpinned pawn of the side attacks two or more higher-value opponent pieces (knight, bishop, rook, queen, or king). \\

\texttt{knight\_fork} [m\,|\,o] &
True if an unpinned knight of the side attacks two or more higher-value opponent pieces (rook, queen, or king). \\

\texttt{bishop\_fork} [m\,|\,o] &
True if an unpinned bishop of the side attacks two or more higher-value opponent pieces (rook, queen, or king). \\

\texttt{rook\_fork} [o] &
True if an unpinned opponent rook attacks two or more higher-value active-player pieces (queen or king). \\

\midrule
\multicolumn{2}{l}{\textit{Tactical Patterns --- Pins \textnormal{(10 concepts)}}} \\
\midrule

\texttt{has\_pinned\_pawn} [m\,|\,o] &
True if the side has a pawn that is pinned to its own king. \\

\texttt{has\_pinned\_knight} [m\,|\,o] &
True if the side has a knight that is pinned to its own king. \\

\texttt{has\_pinned\_bishop} [m\,|\,o] &
True if the side has a bishop that is pinned to its own king. \\

\texttt{has\_pinned\_rook} [m\,|\,o] &
True if the side has a rook that is pinned to its own king. \\

\texttt{has\_pinned\_queen} [m\,|\,o] &
True if the side has a queen that is pinned to its own king. \\

\midrule
\multicolumn{2}{l}{\textit{Positional Structure \textnormal{(23 concepts)}}} \\
\midrule

\texttt{has\_connected\_rooks} [m\,|\,o] &
True if the side has two rooks on the same rank or file with no intervening pieces. \\

\texttt{has\_bishop\_pair} [m\,|\,o] &
True if the side has exactly two bishops. \\

\texttt{has\_control\_of\_\allowbreak open\_file} [m\,|\,o] &
True if the side has a rook or queen on a pawn-free file with no opposing rooks or queens on that file. \\

\texttt{has\_contested\_\allowbreak open\_file} &
True if both sides have a rook or queen on the same pawn-free file. \\

\texttt{capture\_possible\_\{sq\}} [m\,|\,o] \newline sq $\in$ \{d1, d2, d3, e1, e2, e3, g5, b5\} &
True if the side has a legal capture on square \texttt{sq}, evaluated from the side's own board perspective (squares are mirrored for the opponent). Instantiated for each of 8 key squares, yielding 16 concepts. \\

\bottomrule
\end{tabularx}
\end{table}

\Needspace{19\baselineskip}
\subsection{Details for Blundered Transitional Dataset}
\label{app:btd}

\begin{wraptable}{r}{0.47\textwidth}
\centering
\small
\setlength{\tabcolsep}{5pt}
\caption{Dataset construction and current split sizes.}
\label{tab:dataset_stats}
\begin{tabular*}{\linewidth}{@{}l@{\extracolsep{\fill}}r@{}}
\toprule
Candidate transitional positions & 2.28M \\
Blundered transitional positions & 1.15M \\
\midrule
Training split & 0.80M \\
Validation split & 0.11M \\
Test split & 0.23M \\
\bottomrule
\end{tabular*}
\end{wraptable}

We construct the Blundered Transitional Dataset from positions in the Lichess game database that have Stockfish evaluations.\footnote{\url{https://database.lichess.org/\#evals}} First, we run Maia-2 on each position across all 11 skill conditions. Following Section~\ref{sec:data}, we retain positions where its prediction changes monotonically from a suboptimal move at lower skill conditions to the engine-selected best move at higher conditions, yielding 2.28M candidate transitional positions.

Next, we evaluate the lowest-skill prediction with Stockfish at depth 20 and retain positions where it reduces the side-to-move's estimated win probability by more than 10 percentage points relative to the best move~\citep{mcilroy2020aligning}. The current dataset contains 1,147,280 positions, divided into fixed 70/10/20 train, validation, and test splits as shown in Table~\ref{tab:dataset_stats}. Thus, each retained position exhibits a clear behavioral gap while keeping the board and model weights fixed across skill conditions. The concept-specific externalization experiments apply the additional filters described in Appendix~\ref{app:ke_all}.
\WFclear

\subsection{\KEnc{} for All Oracle Chess Concepts}
\label{app:ka_all_concepts}

We train independent linear probes on the average-pooled residual stream after each of Maia-2's two Transformer blocks, separately for each of the 172 oracle concepts and 11 skill levels. For each concept, we collect up to 5,000 positive and 5,000 negative positions, then truncate both classes to the size of the smaller class. We shuffle this balanced dataset with seed 42 and use 70\% for training, 10\% for validation, and 20\% for testing. The positions, labels, and data split remain fixed across skill levels. Each probe uses $\ell_1$ regularization. We select $C \in \{0.01, 0.1, 1, 10, 100\}$ independently for each concept and skill level using validation accuracy, then train the final probe with early stopping and report its test accuracy.

\begin{figure}[!htbp]
    \centering
    \includegraphics[width=0.96\textwidth]{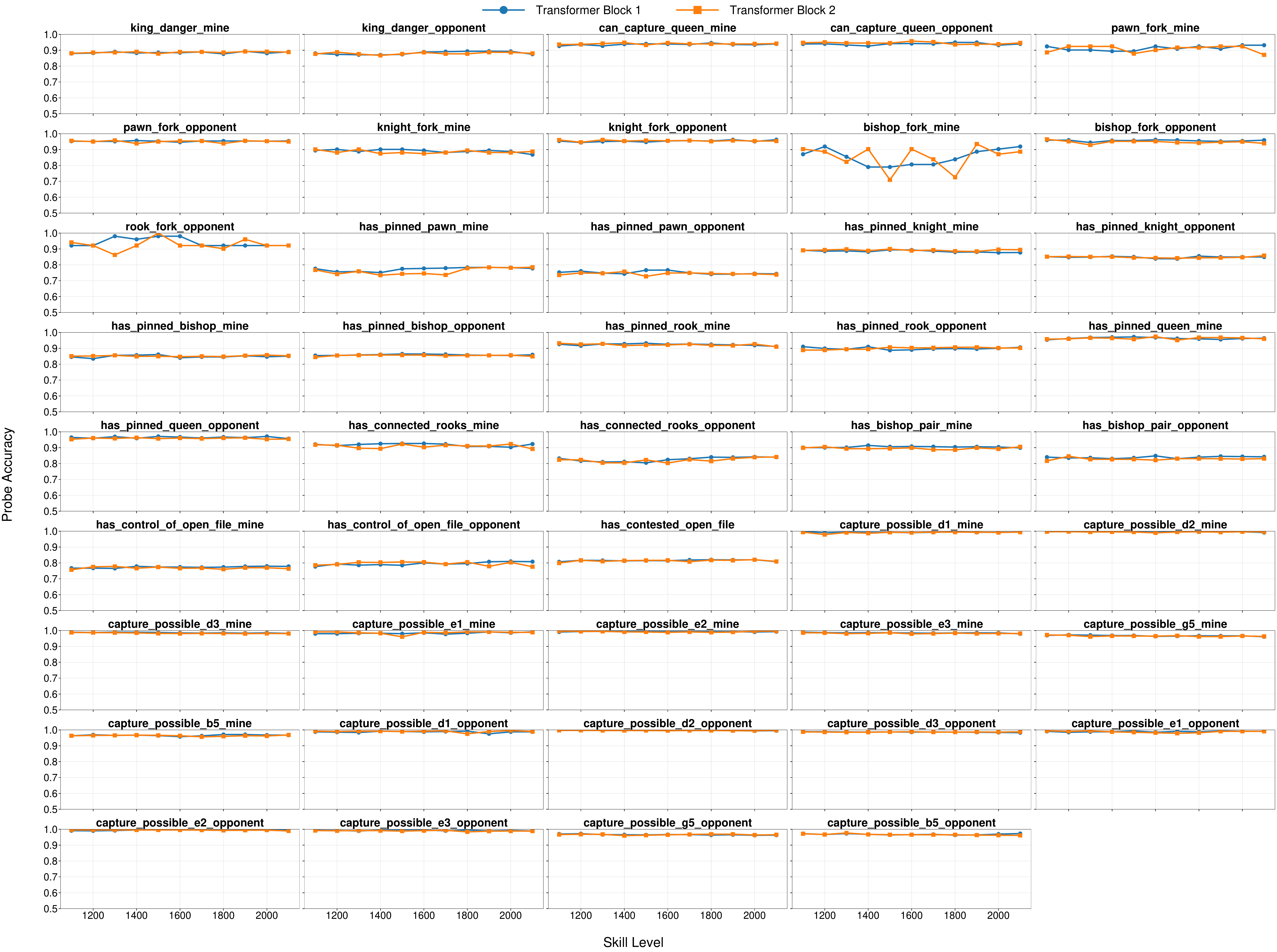}
    \caption{Probe accuracy across skill levels for all 44 king and queen safety, tactical pattern, and positional structure concepts.}
    \label{fig:probes_positional}
\end{figure}

\begin{figure}[!htbp]
    \centering
    \includegraphics[width=0.96\textwidth]{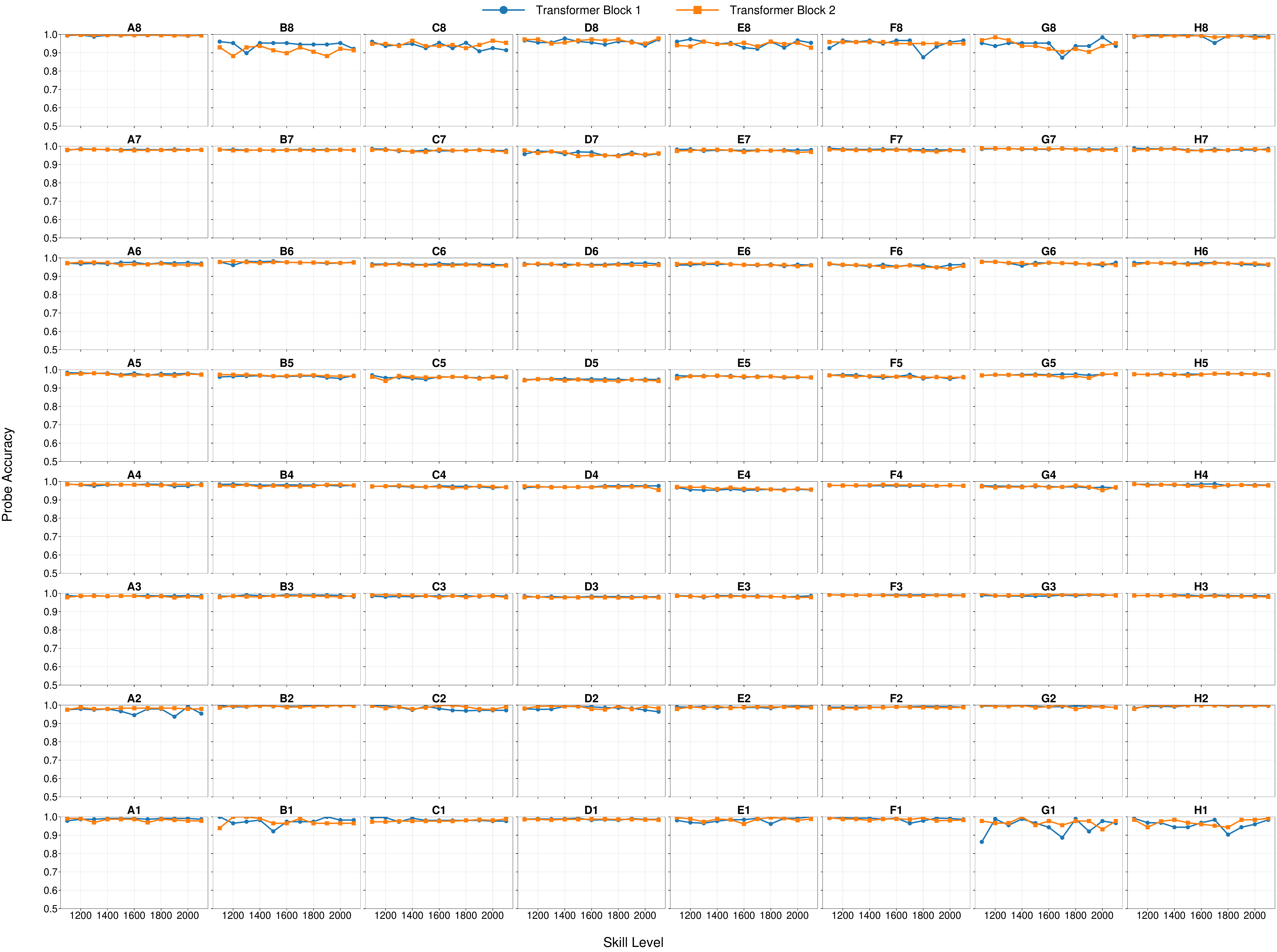}
    \caption{Probe accuracy across skill levels for all 64 per-square control concepts from the active player's perspective.}
    \label{fig:probes_sq_mine}
\end{figure}

\begin{figure}[!htbp]
    \centering
    \includegraphics[width=0.96\textwidth]{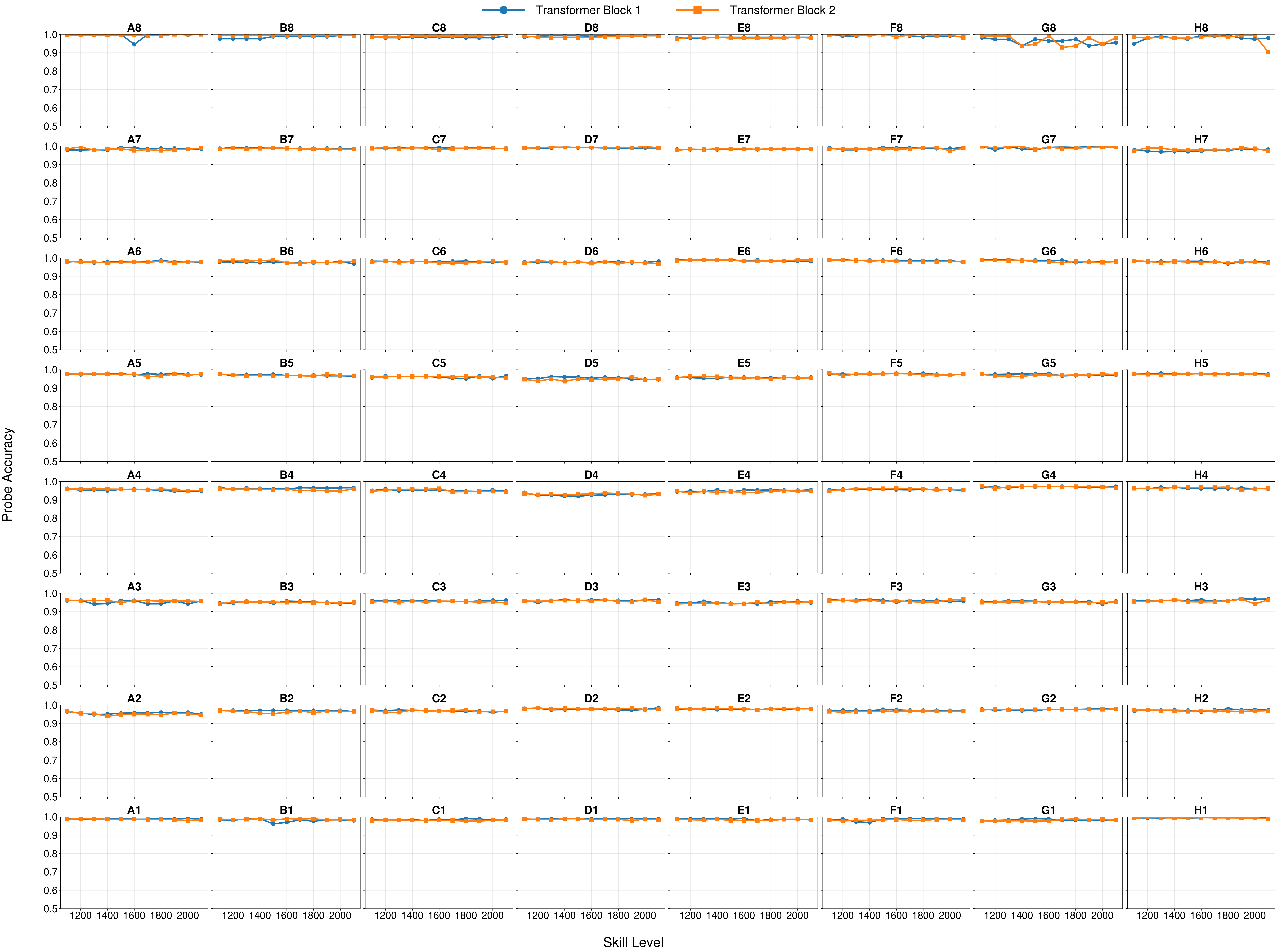}
    \caption{Probe accuracy across skill levels for all 64 per-square control concepts from the opponent's perspective.}
    \label{fig:probes_sq_opponent}
\end{figure}

Figures~\ref{fig:probes_positional}--\ref{fig:probes_sq_opponent} present the complete results. Across all 172 concepts and both Transformer blocks, probe accuracy remains largely flat from the $<1100$ to the $\geq2000$ skill condition. The same pattern appears for king and queen safety, tactical patterns, positional structure, and per-square control. Thus, the aggregate result in Section~\ref{sec:probes} does not conceal a systematic loss of measured concept encoding within a particular category.

\subsection{Cross-Skill Probe Transfer}
\label{app:probe_transfer}

\begin{figure}[!htbp]
    \begin{minipage}[t]{0.48\textwidth}
        \centering
        \vspace{0pt}
        \includegraphics[height=0.74\linewidth,keepaspectratio]{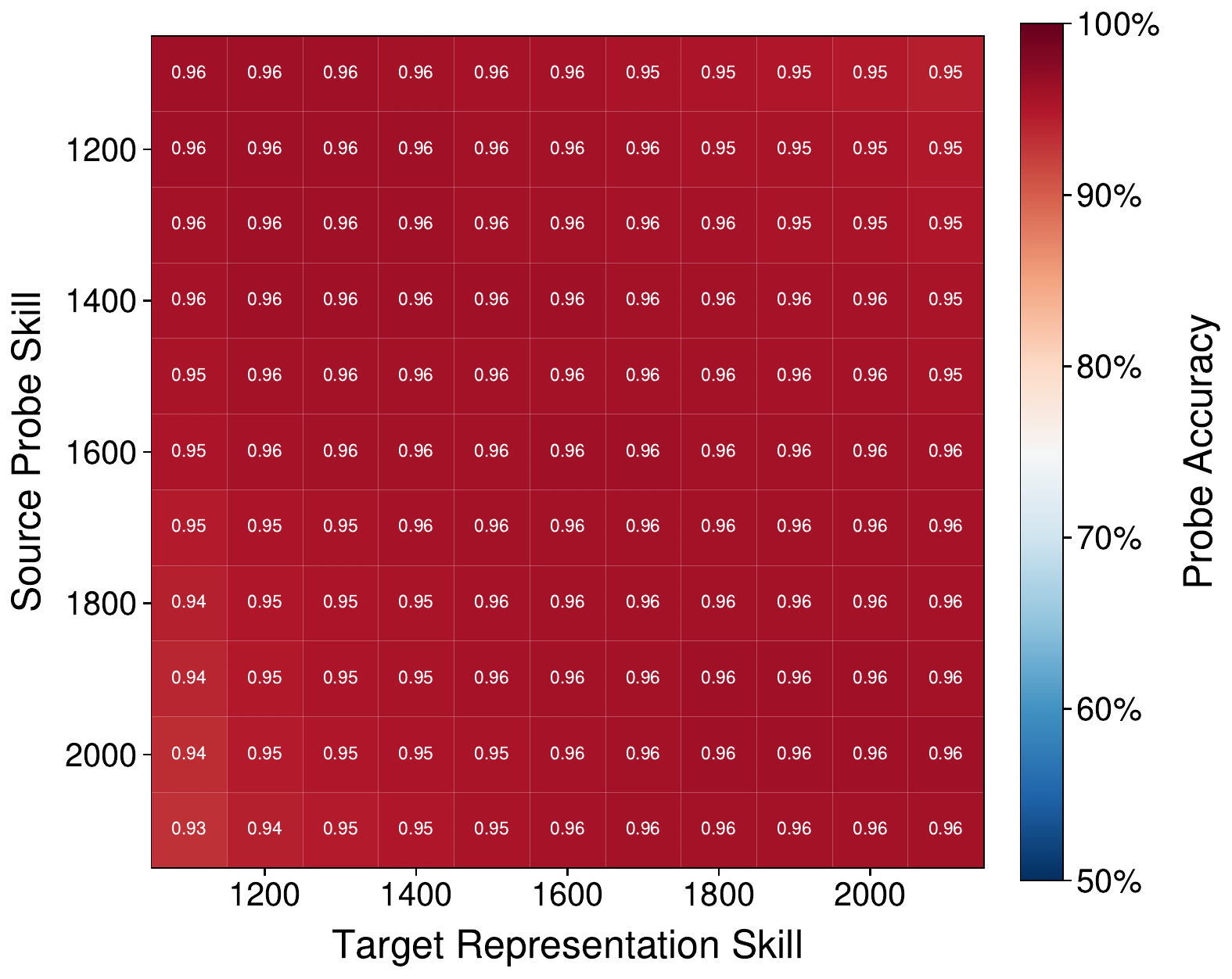}
        \caption{Layer-averaged cross-skill probe accuracy over 172 concepts. Rows denote source-probe skills and columns denote target-representation skills.}
        \label{fig:probe_transfer}
    \end{minipage}
    \hfill
    \begin{minipage}[t]{0.48\textwidth}
        \centering
        \vspace{0pt}
        \includegraphics[height=0.74\linewidth,keepaspectratio]{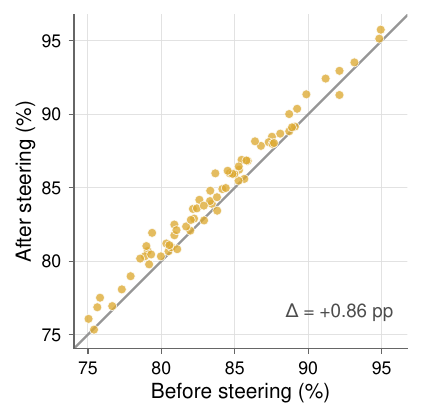}
        \caption{Probe accuracy before and after SAE feature steering. Each point is one concept, averaged over 11 skill levels.}
        \label{fig:sae_encoding_preservation}
    \end{minipage}
\end{figure}

The main probe analysis fits a separate classifier for each concept and skill level. Similar within-skill accuracy could therefore arise even if each skill condition required a different linear readout. To test this possibility, for every source skill $s$ and target skill $t$, we apply the probe $g_s$ without refitting to the same held-out positions represented under $t$. The positions and labels are fixed across target skills and excluded from probe training. This yields $A_{s,t}=\operatorname{Acc}(g_s(\mathbf{z}_t),c)$. We first average each entry equally over the 172 concepts and then average the two Transformer blocks, whose transfer patterns are nearly identical, to produce Figure~\ref{fig:probe_transfer}.

Matched-probe accuracy is 96.0\% for Transformer Block 1 and 95.9\% for Block 2, while mean off-diagonal accuracy is 95.6\% and 95.4\%. The corresponding gaps are 0.40 percentage points (95\% bootstrap CI: 0.34--0.47) and 0.43 points (0.36--0.50). The gap increases smoothly with skill distance: highest-skill probes retain 93.2\% and 92.8\% accuracy on the lowest-skill representations. In the reverse direction, lowest-skill probes reach 94.7\% and 94.3\% on the highest-skill representations. Thus, skill conditioning induces a small, continuous change in representation, but the concept information remains accessible through a largely shared linear readout across the full skill range.

\FloatBarrier
\subsection{Extended Results for Modulating \KExt{}}
\label{app:ke_all}

\begin{figure}[!htbp]
    \centering
    \includegraphics[width=0.98\textwidth,trim=0 5pt 0 0,clip]{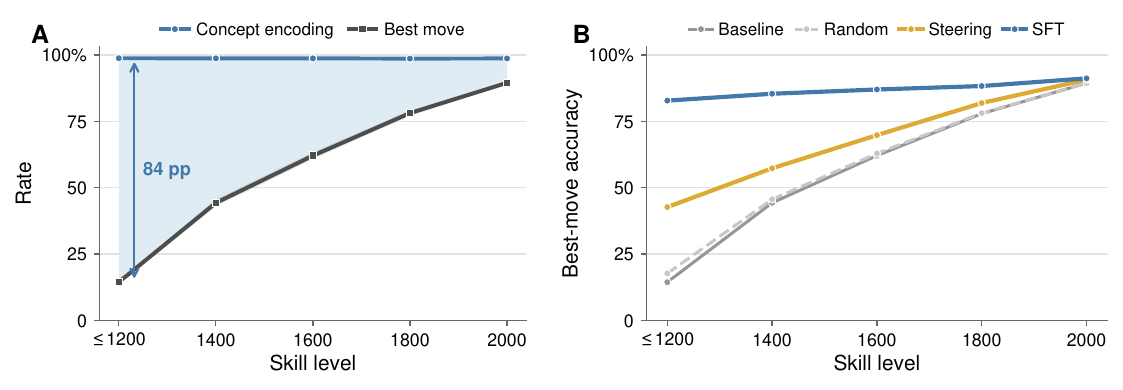}
    \vspace{-9pt}
    \caption{Maia-2's concept encoding and best-move accuracy on the same positions with target concepts (A), and recovery under feature steering and Head SFT across skill levels (B).}
    \label{fig:central_evidence}
\end{figure}

\xhdr{Additional results}
Figure~\ref{fig:central_evidence} compares concept encoding and best-move accuracy on the same positions containing the target concepts. Encoding remains near ceiling as move accuracy changes across skill levels. Both policy-head fine-tuning (Head SFT) and sparse autoencoder (SAE) feature steering narrow the behavioral gap, with Head SFT recovering most of it.

\xhdr{Concept selection for externalization experiments} Of the 172 oracle concepts, not all yield enough qualifying positions for the intervention experiments. A position qualifies for concept $c$ only if: (i) the oracle detects $c$, (ii) the oracle confirms that the optimal move addresses $c$, and (iii) the lowest-skill prediction fails to address $c$ and causes a blunder. Here, a move \emph{addresses} a concept if it takes the corresponding concept-relevant action, such as capturing an under-defended opponent piece or relocating or defending a threatened piece. We require at least 500 qualifying positions per concept to support train, validation, and test splits together with intervention hyperparameter selection. This leaves 71 concepts, and all results in Section~\ref{sec:validate_2} use this set.

\xhdr{Feature-steering site and control} As detailed in Appendix~\ref{app:feature_steering}, SAE steering acts on the mean-pooled FFN contribution in Transformer Block 2 and reinjects the decoded representation before residual addition. Steering (Random) follows the same procedure with uniformly sampled SAE features, controlling for a generic activation perturbation.

\begin{figure}[!htbp]
    \centering
    \begin{overpic}[width=0.99\textwidth]{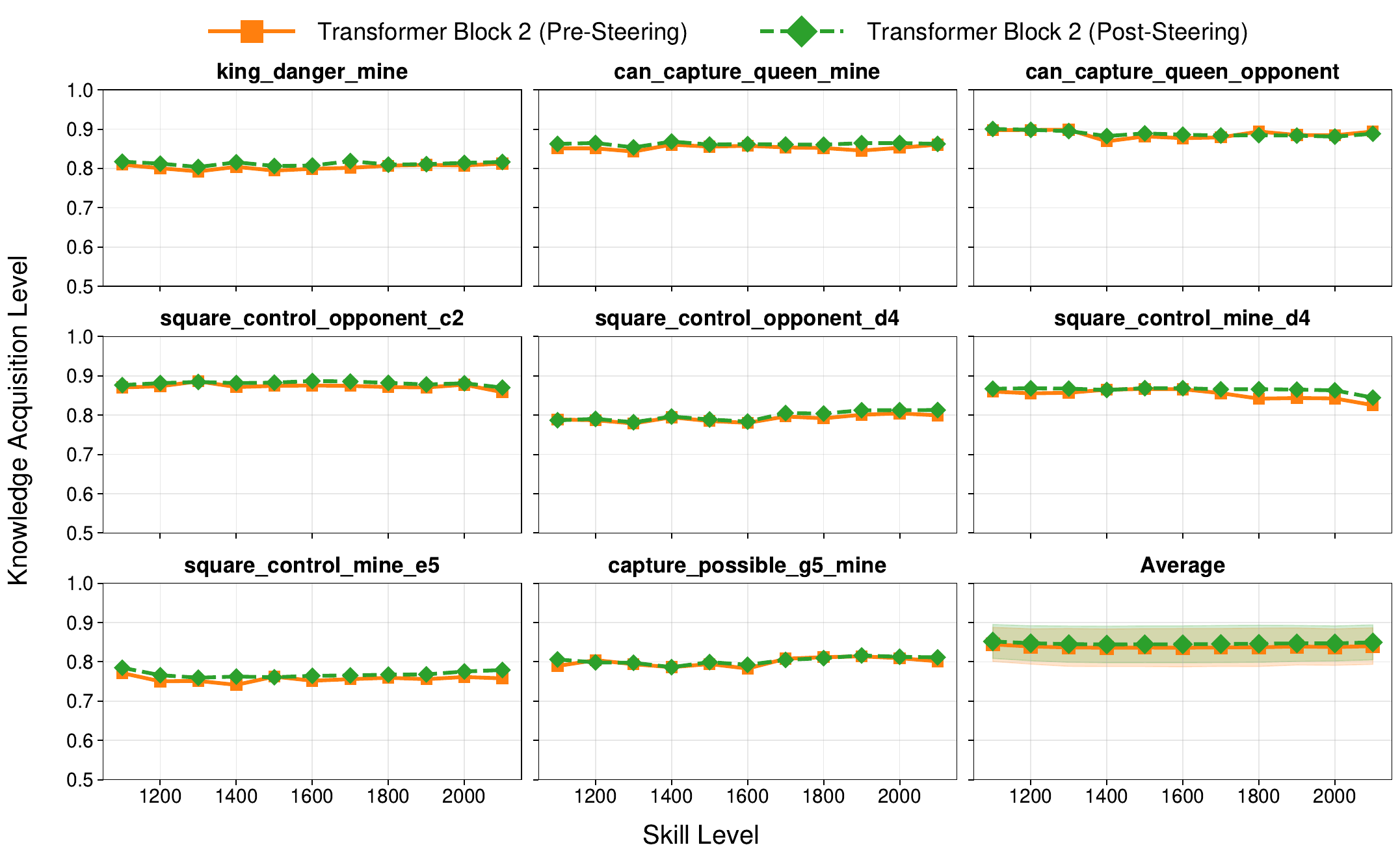}
        \put(0,18){\color{white}\rule{3.2\unitlength}{27\unitlength}}
        \put(0.7,25){\rotatebox{90}{\small Probe Accuracy}}
        \put(10.5,37.1){\color{white}\rule{24\unitlength}{3\unitlength}}
        \put(14.0,37.8){\small\bfseries defensive\_threat\_c2}
        \put(40.5,37.1){\color{white}\rule{26\unitlength}{3\unitlength}}
        \put(44.0,37.8){\small\bfseries defensive\_threat\_d4}
        \put(73.5,37.1){\color{white}\rule{22\unitlength}{3\unitlength}}
        \put(77.0,37.8){\small\bfseries offensive\_threat\_d4}
        \put(11.5,19.5){\color{white}\rule{22\unitlength}{3\unitlength}}
        \put(14.2,20.2){\small\bfseries offensive\_threat\_e5}
    \end{overpic}
    \caption{Probe accuracy before and after SAE feature steering across 11 skill levels for eight representative concepts and the 71-concept average.}
    \label{fig:post_steering_probes}
\end{figure}

\xhdr{\Kenc{} after feature steering} Figure~\ref{fig:sae_encoding_preservation} summarizes \kenc{} before and after SAE feature steering across all 71 concepts, while Figure~\ref{fig:post_steering_probes} shows the corresponding curves for representative concepts. For the statistical comparison, we first average the 11 skill levels within each concept and then pair the pre- and post-steering accuracies across concepts. The mean change is $+0.86$ percentage points (90\% CI: $[0.73,0.98]$), which falls within a $\pm 1$-point equivalence margin under paired two one-sided tests (TOST; $p=0.031$).

\section{Details for SAE Feature Steering}
\label{app:feature_steering}

\subsection{Maia-2 SAE Training}
\label{app:sae}

We train separate Sparse Autoencoders (SAEs) on the FFN contribution from each of Maia-2's two Transformer blocks to obtain sparse feature dictionaries~\citep{cunningham2023sparse,bricken2023sparse}. We then identify features associated with oracle concepts and use the Block~2 dictionary for feature steering.

\xhdr{Training data} We use rated Lichess games from September through November 2023, which were also included in Maia-2's training corpus. We follow Maia-2's balancing and filtering procedure: games are balanced across player skill levels, while the first 10 plies and moves made with less than 30 seconds remaining are excluded~\citep{mcilroy2020aligning}.

\xhdr{Activation collection} For each Transformer block, we collect the FFN contribution before it is added to the residual stream, mean-pool across the eight patch tokens, and train a separate SAE on the resulting 1024-dimensional vectors.

\xhdr{Architecture and objective} We use the JumpReLU architecture~\citep{rajamanoharan2024jumping}, with encoder $W_e \in \mathbb{R}^{d_{\text{hidden}} \times d_{\text{in}}}$ and decoder $W_d \in \mathbb{R}^{d_{\text{in}} \times d_{\text{hidden}}}$. Here, $d_{\text{in}}$ is the FFN output dimension and $d_{\text{hidden}}$ is the dictionary size. We minimize
\begin{align}
\mathcal{H} &= \operatorname{JumpReLU}_{\boldsymbol{\theta}}(W_e\mathbf{x}+\mathbf{b}_e),
& \hat{\mathbf{x}} &= W_d\mathcal{H}+\mathbf{b}_d, \\
\mathcal{L}_{\text{SAE}} &= \|\mathbf{x} - \hat{\mathbf{x}}\|_2^2 + \lambda_{\text{L0}}\|\mathcal{H}\|_0,
\end{align}
where $\boldsymbol{\theta}$ contains the learned activation thresholds and $\lambda_{\text{L0}}$ controls sparsity. We initialize encoder and decoder weights with unit-normalized He initialization~\citep{he2015delving} and set both biases to zero.

\begin{figure}[!htbp]
    \centering
    \includegraphics[width=0.99\textwidth]{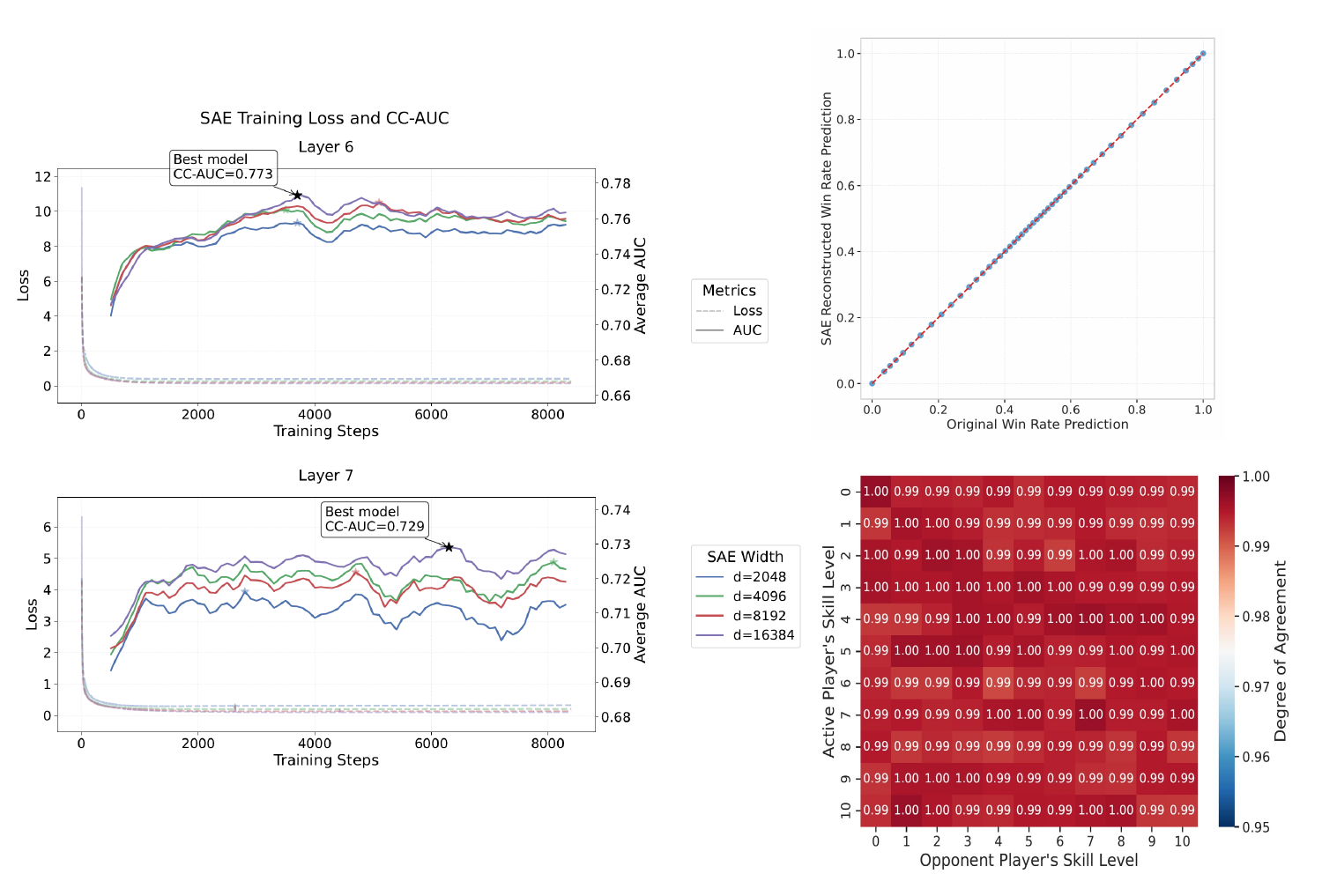}
    \caption{Training diagnostics and predictive fidelity for Maia-2 SAEs.}
    \label{fig:sae_training}
\end{figure}

\xhdr{Model selection} We search dictionary sizes from 1$\times$ to 16$\times$ the FFN output dimension and sparsity coefficients from $5\times10^{-6}$ to $10^{-4}$, training each configuration for up to 10,000 iterations. Because reconstruction loss alone does not show whether a dictionary captures chess structure, we follow \citet{karvonen2024measuring} and track the best feature-level ROC-AUC over 93 labeled chess concepts. This calibration set comprises 29 board-state properties, such as checks, open files, and capture opportunities on key squares, and 64 piece-location indicators, and is separate from the oracle used in Sections~\ref{sec:h1_knowledge_encoding} and~\ref{sec:h2_knowledge_externalization}. We stop after 10 evaluation intervals without improvement and retain the checkpoint with the highest aggregate concept AUC for each block. The selected 16$\times$ dictionaries contain 16,384 features. Finally, Figure~\ref{fig:sae_training} shows that reconstructing through these SAEs closely preserves Maia-2's win-rate predictions and move choices across skill levels.

\subsection{Concept-Salient Feature Identification}
\label{app:feature_selection}

Because each oracle provides deterministic position-level labels, we
can measure feature--concept associations directly~\citep{karvonen2024measuring}.

For each oracle concept $c$, we collect a balanced set of positions
where $c$ is present and absent, run the positions through Maia-2 under
the middle skill condition, and encode the mean-pooled FFN contributions
through the trained SAE to obtain sparse feature activations
$\mathcal{H} \in \mathbb{R}^{d_{\text{hidden}}}$. We then score each
SAE feature $j$ by its discriminative power for concept $c$:
\begin{equation}
r(j, c) = \text{AUC}(j, c),
\end{equation}
where $\text{AUC}(j, c)$ is the ROC-AUC score between feature $j$'s
activations and the concept labels across all collected positions. For each concept, we retain the top 100
features ranked by $r(j, c)$. We compute these rankings separately for both SAE dictionaries and use the Block~2 rankings for steering.

\subsection{Feature Steering}

\label{app:steering}

Given the concept-salient features identified in
Section~\ref{app:feature_selection}, we steer Maia-2's move
predictions by amplifying these features during the forward pass,
motivated by the feature steering methodology of
\citet{templeton2024scaling}.

For a given position, we first run a clean forward pass to obtain the
Block~2 FFN contribution and encode its mean-pooled representation through the SAE to obtain
$\mathcal{H}$. We then amplify the top-$k$ salient features for the
target concept by a factor $\alpha$:
\begin{equation}
\tilde{\mathcal{H}}_j = \begin{cases}
\alpha \cdot \mathcal{H}_j & \text{if } j \in \mathcal{F}_c^k \\
\mathcal{H}_j & \text{otherwise}
\end{cases}
\end{equation}
where $\mathcal{F}_c^k$ denotes the top-$k$ features for concept $c$.
The modified latent representation $\tilde{\mathcal{H}}$ is
reconstructed through the SAE decoder, broadcast across the eight patch
tokens, and substituted for the original Block~2 FFN contribution before residual addition. A second
forward pass produces the steered move predictions.

\xhdr{Hyperparameter selection} We optimize $k$ and $\alpha$ for the ten
non-highest skill conditions via grid search on the training split of the
concept-filtered dataset, with $k \in \{1, 2, 5, 10, 20\}$ and
$\alpha \in \{2, 5, 10, 20\}$, selecting the combination that
maximizes the proportion of predictions matching the optimal move. The
highest skill condition reuses the setting from the adjacent condition.

\xhdr{\KEnc{} after steering} Unlike Head SFT, feature steering modifies
internal activations, so stable \kenc{} is not guaranteed by
construction. We therefore apply the probes trained before steering
(Section~\ref{sec:probing}) to the intervened Block~2 representations.
Appendix~\ref{app:ke_all} reports the resulting probe curves and paired
equivalence test.

\FloatBarrier
\Needspace{0.60\textheight}
\section{Additional Related Work}
\label{app:additional_related}

\xhdr{Oracle chess engines} Chess has long served as a benchmark for AI, from Deep Blue~\citep{hsu1999ibm} to superhuman engines such as Stockfish~\citep{stockfish} and AlphaZero~\citep{silver2018general}. For our purposes, engine evaluations provide a consistent reference for assessing move quality and identifying best moves.

\xhdr{Reading internal representations} Probing is widely used to study knowledge encoding by measuring how accurately a simple predictor can recover a property from a model's internal representations~\citep{alain2016understanding,gurnee2023language,burns2023discovering}. Logit and tuned lenses provide a complementary readout by mapping intermediate activations to distributions over the vocabulary~\citep{nostalgebraist2020logit,belrose2023eliciting}. Readouts that combine signals across internal layers can also outperform methods based on model outputs or final-layer representations~\citep{jiao2023spin,jiao2026siren,jiao2026agent}. This literature motivates our operational measure of \kenc{} as concept decodability, which we distinguish from the use of those concepts in move selection.

\xhdr{Intervening on internal computations} Interventions complement these readouts by testing how changes to internal computations affect model behavior. Sparse autoencoders decompose activations into features that can be amplified or ablated to test their influence on outputs~\citep{bricken2023sparse,templeton2024scaling,zou2026deciphering}. Activation patching and attention-head ablation test the contributions of specific components by replacing their activations or suppressing their outputs~\citep{meng2022locating,wang2022interpretability,jiao2026conflict}. We use policy-head fine-tuning and SAE feature steering to test whether lower-skill representations retain knowledge that can support stronger move predictions.

\end{document}